\pdfoutput=1

\documentclass[11pt]{article}

\usepackage{acl}

\usepackage{times}
\usepackage{latexsym}

\usepackage[T1]{fontenc}

\usepackage[utf8]{inputenc}

\usepackage{microtype}

\usepackage{inconsolata}

\usepackage{tikz}
\usetikzlibrary{positioning}
\usepackage{amsmath}
\usepackage{booktabs}
\usepackage{subcaption}
\usepackage{multirow}
\usepackage{algorithm}
\usepackage{algpseudocode}
\usepackage{algpseudocode}
\usepackage{fdsymbol}

\usepackage{dcolumn}
\usepackage{siunitx}

\usepackage{authblk}

\title{Best of Both Worlds: A Pliable and Generalizable Neuro-Symbolic Approach for Relation Classification}

\author[$\spadesuit$, $\clubsuit$]{Robert Vacareanu}
\author[$\spadesuit$]{Fahmida Alam}
\author[$\spadesuit$]{Md Asiful Islam}
\author[$\spadesuit$]{Haris Riaz}
\author[$\spadesuit$]{Mihai Surdeanu}
\affil[$\spadesuit$]{University of Arizona, Tucson, AZ, USA}
\affil[$\clubsuit$]{Technical University of Cluj-Napoca, Cluj-Napoca, Romania
}
\affil[ ]{\textit {\{rvacareanu,fahmidaalam,asifulislam,hriaz,msurdeanu\}@arizona.edu}}

\begin{document}
\maketitle
\begin{abstract}
This paper introduces a novel neuro-symbolic architecture for relation classification (RC) that combines rule-based methods with contemporary deep learning techniques. 
This approach capitalizes on the strengths of both paradigms: the adaptability of rule-based systems and the generalization power of neural networks. Our architecture consists of two components: a declarative rule-based model for transparent classification and a neural component to enhance rule generalizability through semantic text matching.
Notably, our semantic matcher is trained in an unsupervised domain-agnostic way, solely with synthetic data.
Further, these components are loosely coupled, allowing for rule modifications without retraining the semantic matcher.
In our evaluation, we focused on two few-shot relation classification datasets: Few-Shot TACRED and a Few-Shot version of NYT29. 
We show that our proposed method outperforms previous state-of-the-art models in three out of four settings, despite not seeing any human-annotated training data.
Further, we show that our approach remains modular and pliable, i.e., the corresponding rules can be locally modified to improve the overall model. Human interventions to the rules for the TACRED relation \texttt{org:parents} boost the performance on that relation by as much as 26\% relative improvement, without negatively impacting the other relations, and without retraining the semantic matching component.
\end{abstract}

\section{Introduction}
\label{s:introduction}

After the ``deep learning tsunami'' \cite{Manning2015ComputationalLA}, neural approaches for information extraction (IE) consistently pushed the boundaries of the state of the art \cite{Yang2016HierarchicalAN, zhang2017tacred, guo-etal-2019-attention, Yamada2020LUKEDC, Zhong2020AFE}. However, all these directions come at a cost: (i) low explainability \cite{danilevsky_2021_explainability_for_nlp} and (ii) fragility \cite{Sculley2015HiddenTD}.

\begin{figure}[t]
    \resizebox{\columnwidth}{!}{\begin{tabular}{ll}
    \toprule
    \textbf{\Large{Rule}} & \texttt{[ne=per]+ <nsubj founded >dobj [ne=org]+} \\
    \addlinespace[0.25em]
    \midrule
    \addlinespace[0.25em]
    \textbf{\Large{Sentence 1}} & \Large{Bill Gates founded Microsoft} \\
    \addlinespace[0.5em]
    \textbf{\Large{Sentence 2}} & \Large{Bill Gates is the founder of Microsoft} \\
    \addlinespace[0.5em]
    \textbf{\Large{Sentence 3}} & \Large{John moved to New York City} \\
    \bottomrule
\end{tabular}}
    \vspace{-2mm}
      \caption{\footnotesize{An example of the type of rules we use in our proposed method, together with three sentences. The rule captures the {\tt org:founder} relation with a syntactic pattern anchored by the predicate {\em founded} that has a person named entity as its subject and an organization as the direct object. 
      By itself, the rule matches the first sentence, but it does not match the other two. 
      When coupled with our semantic matching component, the rule matches the first two sentences.
      }}
    \label{fig:rule_example}
     \vspace{-5mm}
\end{figure}

Explainability is critical in many domains such as healthcare, law, and finance \cite{Adadi2018PeekingIT, Goodman2016EuropeanUR, Tjoa2019ASO}. While there have been efforts to incorporate explainability into neural methods \cite[inter alia]{ribeiro_2016_lime, lundberg_2017_shap, Tang2022ItTT}, most explanations are \textit{local} and \textit{post-hoc}, which has two important drawbacks. First, such explanations are not guaranteed to be \textit{faithful} \cite{Jacovi2020TowardsFI}. Second, they are not {\em actionable}. That is, it is not immediately possible to modify the underlying model using insights from the explanations without risking introducing new, unforeseen behavior.
In contrast, rule-based\footnote{We refer to syntactic and surface patterns as rules, such as, \texttt{[ne=per]+ <nsubj founded >dobj [ne=org]+}.} methods are explainable and {\em pliable},\footnote{Term introduced by Dayne Freitag in the panel discussion at the PaN-DL workshop \cite{pandl-2022-pattern} to indicate that rules can be modified to improve the corresponding local behavior while minimizing the impact on the rest of the model. \label{fn:pliability}} but lack the generalization power of current deep learning systems \cite{Tang2022ItTT}. 

In this paper, we propose a novel neuro-symbolic architecture for relation classification (RC) that preserves the advantages of both directions, i.e., the generalization of neural methods and the pliability of rule-based approaches with a modular approach, containing two components: a declarative rule-based model and a neural component. The first module implements relation classification with a set of explainable rules. The second increases the generalizability of rules by semantically matching them over text.
Figure~\ref{fig:rule_example} shows an example of how the two components interact.

Our specific contributions are:
{\flushleft {\bf (1)}} We propose a {\em modular neuro-symbolic architecture for relation classification} that combines the advantages of symbolic and neural models. 
The symbolic rule-based component utilizes syntactic or surface rules automatically derived from example sentences, formulated as the shortest syntactic paths between two entities within a sentence. 
The neural model, which semantically matches these rules over text, is trained {\em without any human-annotated data}. This training involves a unique process: sentences are randomly selected from a large corpus, and rules are automatically generated between random entities in these sentences. The model is then trained in a contrastive manner to assign a high score to the original (rule, sentence) pair (or a paraphrase of the sentence) and a low score otherwise.
The semantic matcher is then combined with the original rule-based model in a two-stage sieve architecture that prioritizes the higher-precision component.

{\flushleft {\bf (2)}} We obtain state-of-the-art performance on three out of four settings in two challenging few-shot RC datasets --Few-Shot TACRED \cite{zhang2017tacred,sabo-etal-2021-revisiting} and a few-shot version of the NYT29 dataset~\cite{riedel2010modeling,takanobu2019hierarchical,Fahmida2024MetaDatasetRE}, without using the background training dataset. For example, on TACRED we observe an improvement of over 12 F1 points over previous state-of-the-art neural-based supervised methods; our overall results on TACRED are $24.52$ for 1-shot and $34.48$ for 5-shot, {\em despite never training the model on any annotated examples from this dataset}. Further, the resulting model is relatively \textit{small}, with approximately 350M parameters.

{\flushleft {\bf (3)}} We show that our approach is {\em pliable} through a user study in which two domain experts manually improved the rules for the {\tt org:parents} relation in TACRED. {\em Without retraining the semantic-matching neural component}, the performance for this relation increases in all settings for both experts, without impacting negatively the performance for the other relations. To our knowledge, this is the first work that shows that pliability can be preserved in neural directions for IE.

\section{Related Work}
\label{s:related_work}

We overview the two main directions that influenced this work --rule-based approaches and explainable deep learning methods-- as well as differences between the proposed work and prompting/in-context learning.

\subsection{Rule-based Approaches}
Rule-based methods were a popular direction for information extraction (IE) before the deep learning era. In the seminal work of \citet{hearst-1992-automatic}, the author proposed a method to learn pairs of words satisfying the hyponymy relation, starting from a simple hand-written rule.
In \citet{Riloff1993AutomaticallyCA}, the author introduced \textit{AutoSlog}, a system capable of learning domain specific relations starting from hand-written patterns. The system was subsequently improved in \citet{Riloff1996AutomaticallyGE} using statistical techniques.
Some approaches towards automatically learning the patterns include \cite{Riloff1999LearningDF, Riloff2003LearningEP, Gupta2014ImprovedPL, vacareanu2022patternrank}; the typical direction is to employ a bootstrapping algorithm, repeatedly alternating between generating rules and generating extractions with the current rules.
Such approaches provided the desired explainability and pliability, but, in retrospect, lacked the generalization capabilities of deep learning methods.


\subsection{Explainable Deep Learning}

Deep learning models have been the preferred approach for the vast majority of NLP tasks including information extraction (IE) in the past years \cite{Hochreiter1997LongSM, sutskever2014seq2seq, vaswani2017attention, devlin19bert}. However, this expressivity came at a cost: numerous articles reported on the fragility of the neural networks \cite{Szegedy2014IntriguingPO, Ilyas2019AdversarialEA, McCoy2019RightFT}, and that neural networks can reinforce biases in the data \cite{Bolukbasi2016ManIT,Brunet2019UnderstandingTO,Mehrabi2021ASO}. As such, having an explainable system is desirable, as long as it does not come at a high cost with respect to performance. The popular approaches to explaining neural networks are either: (i) feature importance, or (ii) surrogate models \cite{danilevsky_2021_explainability_for_nlp}. 

Techniques based on feature importance aim to highlight the feature responsible for a given prediction. For example, \citet{Sundararajan2017AxiomaticAF} uses integrated gradients to assign an importance score to each feature. Other techniques use the attention mechanism as an explanation of the model's prediction \cite{Bahdanau2015NeuralMT, Xu2015ShowAA}. Such techniques show that a feature is important, but do not show how it is being used in the model. Moreover, techniques such as interpreting attention scores have been shown to be particularly brittle. For example, \citet{jain19attentionisnot} has shown that many seemingly different attention patterns can allow for the same end prediction, which raises the question of explanation fidelity. Other improved attention interpretation methods include \citet{Kobayashi2020AttentionIN}, which suggest taking the norm of the vectors into consideration as well.

Techniques based on surrogate models train a (typically) smaller and more interpretable model to explain the original one. For example, \citet{ribeiro_2016_lime} train a linear classifier around the point that is to be explained. \citet{lundberg_2017_shap} uses SHAP values as a unified measure of feature importance. SHAP values are Shapley values \cite{Shapley1988AVF} of a conditional expectation function of the original model. The key issue with surrogate models is their potential lack of fidelity with respect to the original model \cite{danilevsky_2021_explainability_for_nlp}.

\citet{Zhou2020NEROAN} proposed an approach in the same space to ours, i.e., they also train a semantic (or ``soft'') rule matcher (SRM). However, there are multiple critical differences from our work. First, the SRM is used only to augment the training data for a ``traditional'' opaque deep learning RC model, which is the actual output of the training process. In our approach, the SRM is a critical component of the model used during inference.
Second, their SRM module was developed only for surface rules consisting of word constraints, and it is unclear how to expand it to more general patterns.\footnote{For example, their model cannot accommodate more expressive rules that use syntax such as \texttt{[ne=per]+ <nsubj founded >dobj [ne=org]+}.} In contrast, the rules we use in our proposed method are closer to real-world application, i.e., they contain syntactic dependency constraints and semantic entity constraints. Furthermore, their proposed approach requires an initial set of labeled data, while we operate solely in a zero-shot fashion.

All in all, while both (i) feature importance and (ii) surrogate models can provide insights into how and why the deep learning model makes a certain prediction, they do not provide any systematic mechanism to make interventions to these systems.

\subsection{Prompting and In-context Learning}
Lastly, we note that, despite superficial similarities, our work is considerably different from prompting and in-context learning~\cite{brown2020language,schick2020exploiting}. Unlike prompts, our rules are an integral part of the model, both explicitly and through the rule representations learned by our semantic rule matching component. 
Further, rules offer a higher degree of expressiveness compared to raw text. Rules allow humans to unambiguously compress abstract concepts (e.g., by incorporating syntax and semantics) towards a specific goal. In contrast, with prompting and in-context learning, the level of generalization and abstraction is uncertain \cite{lu2021fantastically}.

As we show in Section~\ref{s:experiments}, these advantages make our method obtain state-of-the-art (SOTA) performance as well as more controllable/pliable behavior. Further, in-context learning tends to perform well only with large language models. In contrast, our neural component uses a much smaller language model, containing, with approximately 350M parameters. 

\section{Proposed Method}
\label{s:method}

We propose a hybrid model that combines the advantages of rule-based and neural approaches. Our approach first attempts to strictly match rules, i.e., all semantic/syntactic/lexical constraints must match in the input sentence for a match to be considered. If no rule matches, we back off to a neural semantic rule matching (SRM) component that semantically aligns rules with text.

A key aspect of our proposed approach is that we do {\em not} incorporate a \texttt{no\_relation} classifier in any form, such as a NAV or MNAV \cite{sabo-etal-2021-revisiting}. 
This is important as training multiple representation vectors to capture the entire \texttt{no\_relation} space, as proposed in \cite{sabo-etal-2021-revisiting} can be difficult in practice, as reported by the original authors. 
Instead, our method is simpler: we have rules with associated underlying relations and a single threshold $t \in [0, 1]$ to decide whether the SRM assigned score between a rule and a sentence constitutes a match or not. This threshold is application-specific and can be selected on a development set.

\subsection{Strict Rule Matching Component}
To implement strict rule matching in our hybrid method we use Odinson \cite{valenzuela-escarcega-etal-2020-odinson}. Odinson is a rule-based IE framework with two key advantages. First, it has the capability to combine surface information with syntactic dependency constraints to create a more expressive rule set. Second, the Odinson runtime engine is optimized for speed, and capable of executing rules consisting of surface and syntactic constraints in near real-time. 
We provide an example of the rules we use in Figure~\ref{fig:rule_example}, together with three sentences, one where the rule matches (Sentence 1) and two where it does not (Sentence 2 and 3), according to the strict matching algorithm in Odinson.
This example highlights the key limitation of traditional rule engines: even though the second sentence is semantically similar to the first, Odinson does not match it because its syntax does not align with the syntactic constraints in the rule.
These are precisely the types of problems we aim to address.

Lastly, we emphasize that our proposed method can work with different rule engines.

\begin{figure}[t]
    \resizebox{\columnwidth}{!}{\begin{tabular}{ll}
    \toprule
    \textbf{\Large{Sentence}} & \includegraphics[width=0.7\textwidth]{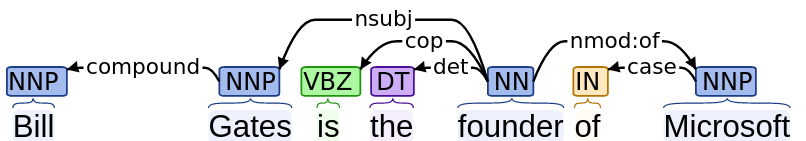} \\
    \addlinespace[0.25em]
    \midrule
    \addlinespace[0.25em]
    \textbf{\Large{Subject Entity}} & \Large{Bill Gates} \\
    \addlinespace[0.5em]
    \textbf{\Large{Object Entity}} & \Large{Microsoft} \\
    \addlinespace[0.5em]
    \textbf{\Large{Relation}} & \Large{org:founder} \\
    \addlinespace[0.25em]
    \midrule
    \addlinespace[0.25em]
    \textbf{\Large{Rule}} & \texttt{[ne=per]+ <nsubj founder >nmod\_of [ne=org]+} \\
    \bottomrule
\end{tabular}

}
      \caption{\footnotesize{To create a rule from a sentence, the process involves: (a) parsing the sentence to extract its syntactic dependency tree, (b) identifying the shortest path connecting two entity mentions within this tree, and (c) constructing a rule based on the syntactic dependencies, associated words, and named entity labels found along this path. For example, the rule shown operates as follows: it requires a \texttt{per} (person) label connected to the word 'founder' via a nominal subject dependency, and 'founder' in turn linked to a \texttt{org} (organization) label through an \texttt{nmod\_of} dependency.}
      }
    \label{fig:rule_generation_example}
     \vspace{-5mm}
\end{figure}

\subsubsection{Rule Generation}
In this paper, we use a simple strategy to generate rules for this component: for syntactic rules, we construct rules from the shortest path in the syntactic dependency tree that connects two entities in a training sentence. For surface rules, we simply take the words in-between the entities. Figure~\ref{fig:rule_generation_example} shows an example of this process. Because we evaluate in a few-shot setting, the number of rules produced for a given relation label will be small, e.g., 1 or 5.

\subsection{Semantic Rule Matching Component}
The example in Figure~\ref{fig:rule_example} highlights the need for a more nuanced approach to rule-based relation classification, one that allows for degrees of matching to overcome the collapse of every non-match to $0$.
To this end, we propose a transformer-based architecture \cite{vaswani2017attention, Liu2019RoBERTaAR, Radford2021LearningTV} that embeds the rule and the sentence; the networks is trained to maximize the cosine similarity between these two embeddings in the case of real matches and minimize it otherwise. We describe the training procedure of our proposed semantic rule matcher below.

\subsubsection{Training Dataset}
\label{method:sss:training_dataset}
A key question is how to obtain training data for the semantic rule matching component, i.e., data that aligns rules with sentences where they should match.
Our method circumvents the need for gold-annotated data, capitalizing on a key insight: for any pair of entities within a sentence, a representative rule can be automatically formulated. Take, for instance, the sentence \textit{John moved to New York City}, featuring entities \textit{John} and \textit{New York City}. From this, we can derive a rule, such as \texttt{[ne=per]+ <nsubj moved >nmod\_to [ne=loc]+} using the underlying syntactic structure of the sentence. This rule, inherently, is indicative of the relationship between these entities, irrespective of the specific nature of this relationship. By applying this principle, we can train our model to assign a high matching score to the tuple consisting this rule and the original entities within their context, while assigning low scores to any other combinations. This innovative approach allows us to automatically create a training dataset, bypassing the traditional reliance on pre-labeled data. 

To encourage the SRM to look beyond syntactic/surface structures, we create paraphrases for the extracted sentences. For example, \textit{John moved to New York City} can be rephrased as \textit{John relocated to New York City} without losing any semantic information. We use this insight to expand the resulting dataset with paraphrases that contain the two entities of interest.\footnote{We use OpenAI's ChatGPT for this purpose (\texttt{gpt-3.5-turbo-1106})} We provide more details below.

We start from UMBC, a dataset of English paragraphs, totaling 3 billion words \cite{han2013umbc}. We pre-process this dataset with standard NLP tools \cite{Manning2014TheSC} for named entity annotations and for dependency parsing. 
Then, we randomly sample a sentence $s_1$ containing two random entities of interest $(e_1, e_2)$, and automatically construct a rule $r_1$ that will match it. The resulting tuple $(r_1, s_1)$ will then be added in the resulting dataset.\footnote{We provide details of the entities we sample in the Appendix~\ref{appendix:entity_types_training_dataset}.}
This process resulted in an initial dataset of approximately 140 million sentence/rule tuples. This dataset is further preprocessed as follows:

\textbf{(1)} We filter the data by removing duplicates and by sub-sampling frequent rules and entities. The underlying motivation is to prevent the model from overfitting to very common rules or entity types. 
For example, the pair \texttt{(ORG, COUNTRY)} is roughly 2 orders of magnitude more common than \texttt{(ORG, EMAIL)}.
At the end of this stage, the resulting dataset has approximately 4 million examples.

\textbf{(2)} We augment the entity types with synonyms, with the goal of encouraging the SRM component to generalize beyond the superficial clues from the entity types. For example, we randomly replace the entity type \texttt{per} with \texttt{human}, or \texttt{individual}. We provide a complete list of the synonyms we used in Appendix~\ref{appendix:entity_types_synonyms}. 

\textbf{(3)} We generate paraphrases of the original sentence, while keeping the two entities of interest in the sentence. 
We use OpenAI's ChatGPT (\texttt{gpt-3.5-turbo-1106}) as our paraphraser, using a simple prompt (shown in Appendix~\ref{appendix:paraphrasing_prompt}). Out of the paraphrases generated, we keep only those that contain the two entities of interest.

In Section~\ref{s:experiments}, we ablate over these three techniques to assess their contribution to the performance of the final model.
In total, the resulting dataset has a total of approximately 5.6 million (rule, sentence) pairs, out of which about 1.6 million pairs were generated through paraphrasing.
When learning sentence representations, we follow prior works on relation extraction \cite{zhang2017tacred, joshi-etal-2020-spanbert, zhou2021improved} and wrap the entities with special tokens, together with the corresponding named entity.
For example, given the entities \textit{Bill Gates} and \textit{Microsoft}, the sentence \textit{Bill Gates founded Microsoft} becomes: \textit{\# * per * Bill Gates \# founded \# * org * Microsoft \#}.

\subsection{Training the Semantic Rule Matching}
\label{sec:train}
\begin{figure}[t]
   \begin{center}
   \includegraphics[width=1.0\columnwidth]{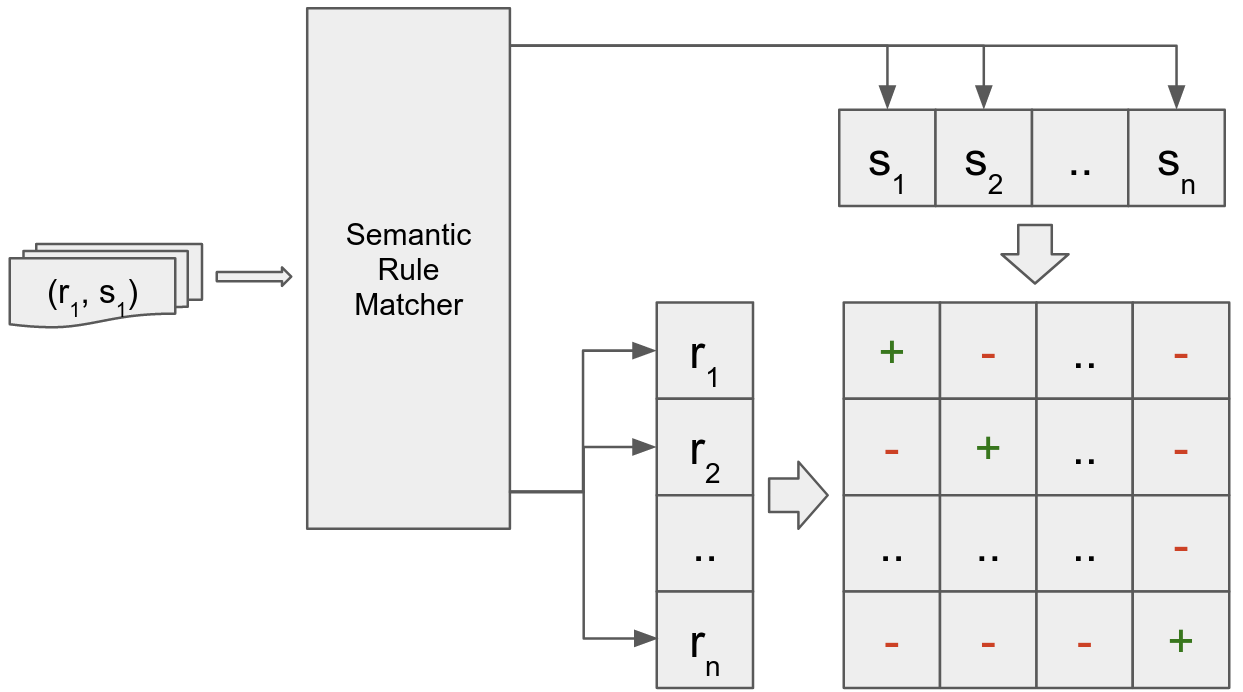}
   \end{center}
   \vspace{-2.5mm}
  \caption{
   \footnotesize{In our training for the Semantic Rule Matcher (SRM), we encode both rules and sentences, followed by calculating cosine similarity between each pair. The goal is to maximize similarity for matching pairs (diagonal of the matrix) and minimize it for non-matching pairs (off-diagonal elements).}
   }
   \label{fig:training_procedure}
   \vspace{-5mm}
\end{figure}

We leverage the resulting dataset to train the semantic rule matching component with a CLIP-like objective. Concretely, the dataset consists of examples of the form $(r, s)$, for example: (\texttt{[ne=per]+ <nsubj founded >nmod\_in [ne=org]+}, \textit{\# * per * Bill Gates \# founded \# * org * Microsoft \#}). We train the SRM component to assign a high cosine similarity score between the embedding of $r$ and the embedding of $s$, and we use the other in-batch examples as negatives \cite{Radford2021LearningTV}. Importantly, {\em we do not use any human-annotated data or any domain-specific relation labels} for training.
We provide an overview of the training mechanism in Figure~\ref{fig:training_procedure}. We use the SRM to encode the rules and the sentences in the current batch. Then, we compute the cosine similarity between every rule and every sentence. Our training objective is then to maximize the similarity scores of matching pairs, found along the diagonal of this matrix. Simultaneously, we minimize the scores of non-matching pairs, which constitute the off-diagonal elements.
Due to space constraints, we include examples of sentences, rules, and their resulting similarities in Appendix~\ref{appendix:qualitative_examples}.

\section{Experiments}
\label{s:experiments}

\subsection{Experimental Setup}
We evaluate our proposed method on Few-Shot TACRED \cite{sabo-etal-2021-revisiting}, a few-shot variant of the TACRED dataset \cite{zhang2017tacred} and on a few-shot variant of the NYT29 dataset~\cite{riedel2010modeling,takanobu2019hierarchical, Fahmida2024MetaDatasetRE}. 
In few-shot settings, the training and testing relation labels are disjoint. We have access to a background training set for tuning the model, but we emphasize that our proposed method does not use it.
Each test sentence is accompanied by 1 (1-shot) or 5 (5-shot) support sentences. 

We use \texttt{RoBERTa-large} \cite{Liu2019RoBERTaAR} for our semantic matching component. Similar to CLIP \cite{Radford2021LearningTV}, we use one model for encoding the rule and one model for encoding the sentence.
We generate rules from the support sentences in each dataset. We use {\tt CoreNLP} \cite{Manning2014TheSC} to obtain the underlying syntactic structure for rule construction. 

At prediction time, we use the proposed method in three ablative configurations: (1) Simply apply the resulting rules in a binary matching fashion, i.e., no SRM (\textbf{Hard-matching Rules}); (2) Use the semantic rule matching module to compute a similarity score between each rule and each sentence, interpreting a similarity above a threshold $t$ as a match\footnote{We tune the threshold on the development partition of each dataset; we do not train on any data from the datastes} (\textbf{Soft-matching Rules}); (3) A combination of (1) and (2), where we first attempt to apply the rules in a typical binary match/no match way (i.e., ``hard'' matching), and if no rule matches we fall back to the semantic rule matching component (i.e. ``soft'' matching). We call this approach \textbf{Hybrid}.

\subsection{Baselines}
We compare our proposed approach with one strong unsupervised baseline and several state-of-the-art supervised approaches from previous work. 

\textbf{Unsupervised Baseline:} Similar to the baseline introduced in \cite{vacareanu2022odinsynth}, this baseline utilizes entity types from query and support sentences for classification, defaulting to \texttt{no\_relation} if no matching types are found.

\textbf{Sentence-Pair:} Employs a transformer-based model to classify concatenated query and support sentences \cite{Gao2019FewRel2T}. We reimplemented this baseline using sentence transformers \cite{reimers-2019-sentence-bert}.

\textbf{MNAV \cite{sabo-etal-2021-revisiting}:} A transformer-based relation classifier is trained on a background set to align vector representations for sentences with identical relations, including multiple vectors for the \texttt{no\_relation} class. During testing, it calculates similarity scores between the test sentence and both the \texttt{no\_relation} vectors and support sentence vectors for each relation. For multiple support sentences of the same relation, it uses an averaged vector representation. The final prediction corresponds to the relation with the highest similarity score.

\textbf{OdinSynth \cite{vacareanu2022odinsynth}:} Utilizes transformer-based rule synthesis from support sentences, predicting the relation with the most rule matches, or \texttt{no\_relation} if there are none.

\begin{table*}
    \resizebox{\textwidth}{!}{\begin{tabular}{@{\extracolsep{4pt}}lrrrrrrr}
  \toprule   
  {Model} & \multicolumn{3}{c}{5-way 1-shot}  & \multicolumn{3}{c}{5-way 5-shot} & \multicolumn{1}{c}{Uses Bacgkround Data}\\
   \cmidrule{2-4} 
   \cmidrule{5-7} 
   \cmidrule{8-8} 
  & \multicolumn{1}{c}{P} & \multicolumn{1}{c}{R} & \multicolumn{1}{c}{F1} & \multicolumn{1}{c}{P} & \multicolumn{1}{c}{R} & \multicolumn{1}{c}{F1} &  \\ 
  \midrule
  Unsupervised Baseline & 5.70 $\pm$ 0.10 & 91.02 $\pm$ 0.65 & 10.73 $\pm$ 0.18     & 5.65 $\pm$ 0.11 & 95.56 $\pm$ 0.70 & 10.67 $\pm$ 0.20 & No \\
  \midrule
  Sentence-Pair \footnotesize{(not fine-tuned)} & 3.9 $\pm$ 0.21 & 5.21 $\pm$ 0.31 & 4.45 $\pm$ 0.24 & 2.76 $\pm$ 0.16 & 8.79 $\pm$ 0.58 & 4.2 $\pm$ 0.25 & No \\
  Sentence-Pair \footnotesize{(fine-tuned)} & 6.89 $\pm$ 0.33 & 28.56 $\pm$ 1.67 & 11.10 $\pm$ 0.55 & 14.94 $\pm$ 0.26 & 24.03 $\pm$ 0.32 & 18.42 $\pm$ 0.16 & Yes \\
  MNAV \footnotesize{(reported)} & - & - & 12.39 $\pm$ 1.01 & - & - & 30.04 $\pm$ 1.92  & Yes \\
  MNAV \footnotesize{(re-run by us)} & 15.11 $\pm$ 0.46 & 8.47 $\pm$ 0.31 & 10.85 $\pm$ 0.29 & 24.48 $\pm$ 1.02 & 32.00 $\pm$ 1.07  & 27.73 $\pm$ 0.94 & Yes \\

  \midrule
  Odinsynth     & 23.48 $\pm$ 1.46 & 11.46 $\pm$ 1.02 & 15.40 $\pm$ 1.21 & 29.77 $\pm$ 0.83 & 20.34 $\pm$ 0.53 & 24.16 $\pm$ 0.44  & No \\
  \midrule
  Hard-matching Rules (ours)  & 51.35 $\pm$ 6.53 & 2.94 $\pm$ 0.48 & 5.56 $\pm$ 0.90           & 45.94 $\pm$ 5.31 & 10.81 $\pm$ 1.23 & 17.50 $\pm$ 1.98 & No \\
  Soft-matching Rules (ours)               & 33.46 $\pm$ 1.47 & 19.69 $\pm$ 1.14 & \textbf{24.78 $\pm$ 1.22} & 51.66 $\pm$ 1.85 & 26.02 $\pm$ 1.29 & \textbf{34.59 $\pm$ 1.24} & No \\
  Hybrid              (ours)  & 32.45 $\pm$ 1.28 & 19.72 $\pm$ 1.08 & 24.52 $\pm$ 1.11 & 44.73 $\pm$ 1.64 & 28.09 $\pm$ 1.40 & 34.48 $\pm$ 1.21 & No \\

  \bottomrule
\end{tabular}

}
      \caption{\footnotesize{The results for the 5-way 1-shot and 5-way 5-shot settings on the test partition of the Few-Shot TACRED dataset. We split the table into 4 blocks as follows:
      (1) a strong unsupervised baseline where the classification is performed based on the types of the entities, (2) state-of-the-art neural methods, (3) rule synthesis using transformer networks, and (4) our proposed method. Our proposed method outperforms previous state-of-the-art methods on both 1-shot and 5-shot splits.}}
    \label{tab:tacred_main_results}
    \vspace{-2mm}
  \end{table*}

\subsection{Main Results}
We present our main results in Tables~\ref{tab:tacred_main_results}~and~\ref{tab:nyt29_main_results} for the standard 1-shot and 5-shot settings on the two datasets. Additionally, we differentiate between methods that use the background training dataset from the ones that do not (i.e., are Zero-Shot).\footnote{By zero-shot we mean methods that do not use human-annotated examples for training.}

We concentrate our discussion on comparing between contemporary rule-based methods (OdinSynth) and strong neural-based methods (MNAV). We draw the following observations.
First, compared to MNAV, the state-of-the-art neural-based method on Few-Shot TACRED, our proposed approach outperforms it in three out of the four settings investigated. For example, in the 1-shot case of Few-Shot TACRED, our proposed method improves upon MNAV by over 12 F1 points (approximately 100\% relative improvement), despite not being trained with any human-annotated data.
Second, compared with Odinsynth, the current best rule-based method on Few-Shot TACRED, our proposed hybrid approach largely outperforms it in both 1-shot and 5-shot cases. This validates the hypothesis that combining a neural network with traditional rule-based approaches outperforms rule-only methods. Importantly, the increased performance does not come at a cost in terms of precision, as our proposed method largely outperforms Odinsynth in terms of precision as well as recall. A similar conclusion holds for the few-shot variant of NYT29 as well.

All in all, our proposed method obtains state-of-the-art performance {\em despite not being trained on any of the human-annotated examples from the respective training datasets}.

\begin{table*}[ht]
    \resizebox{\textwidth}{!}{\begin{tabular}{@{\extracolsep{4pt}}lrrrrrrr}
  \toprule   
  {Model} & \multicolumn{3}{c}{5-way 1-shot}  & \multicolumn{3}{c}{5-way 5-shot} & \multicolumn{1}{c}{Uses Background Data}\\
   \cmidrule{2-4} 
   \cmidrule{5-7} 
   \cmidrule{8-8} 
  & \multicolumn{1}{c}{P} & \multicolumn{1}{c}{R} & \multicolumn{1}{c}{F1} & \multicolumn{1}{c}{P} & \multicolumn{1}{c}{R} & \multicolumn{1}{c}{F1} &  \\ 
  \midrule
Unsupervised Baseline & 11.60 $\pm$ 0.18 &40.34 $\pm$ 0.54 & 18.03 $\pm$ 0.26     & 11.70 $\pm$ 0.25 & 40.65 $\pm$ 0.45 & 18.17 $\pm$ 0.34 & No \\
\midrule
Sentence-Pair \footnotesize{(not fine-tuned)} & 10.61 $\pm$ 0.32 & 12.39 $\pm$ 0.41 & 11.43 $\pm$ 0.35 & 15.81 $\pm$ 0.94 & 5.41 $\pm$ 0.25 & 8.06 $\pm$ 0.39 & No \\
Sentence-Pair \footnotesize{(fine-tuned)} & 38.09 $\pm$ 2.42 & 7.4 $\pm$ 0.42 & 12.4 $\pm$ 0.71 & 36.48 $\pm$ 1.37 & 16.02 $\pm$ 0.41 & 22.26 $\pm$ 0.62 & Yes \\
MNAV          & 25.08 $\pm$ 0.73& 34.37  $\pm$ 0.87 & \textbf{29.00 $\pm$ 0.80} & 33.24 $\pm$ 1.06& 15.47 $\pm$ 0.38  &  21.12 $\pm$ 0.55 & Yes \\
\midrule
OdinSynth     & 30.07 $\pm$ 0.93 & 9.42 $\pm$ 0.31 & 14.34 $\pm$ 0.46 & 21.61 $\pm$ 0.61 & 17.98 $\pm$ 0.45 & 19.63 $\pm$ 0.51 & No \\
\midrule

Hard-matching Rules (ours)  & 77.47 $\pm$ 1.53 & 1.53 $\pm$ 0.13 & 3.01 $\pm$ 0.25           & 80.49 $\pm$ 1.73 & 3.40 $\pm$ 0.12 & 6.52 $\pm$ 0.23 & No \\
Soft-matching Rules (ours)  & 20.80 $\pm$ 0.38 & 12.27 $\pm$ 0.39 & 15.44 $\pm$ 0.40 & 24.50 $\pm$ 0.83 & 16.67 $\pm$ 0.49 & 19.84 $\pm$ 0.59 & No \\
Hybrid              (ours)  & 22.23 $\pm$ 0.47 & 13.45 $\pm$ 0.38 & 16.76 $\pm$ 0.41 & 27.29 $\pm$ 0.77 & 19.52 $\pm$ 0.49 & \textbf{22.76 $\pm$ 0.56} & No \\

  \bottomrule
\end{tabular}

}
      \caption{\footnotesize{The results for the 5-way 1-shot and 5-way 5-shot settings on the test partition of the Few-Shot NYT29 dataset. We split the table into 4 blocks as follows:
      (1) a strong unsupervised baseline where the classification is performed based on the types of the entities, (2) state-of-the-art neural methods, (3) rule synthesis using transformer networks, and (4) our proposed method. Our proposed method obtains the best performance in the 5-shot case, outperforming neural-based methods trained on the background training data.}}
    \vspace{-4mm}
    \label{tab:nyt29_main_results}
  \end{table*}

\subsection{Ablation Analysis}

\begin{table*}[t]
\footnotesize
\resizebox{\textwidth}{!}{\begin{tabular}{rrrrrrrr}
\toprule
 &   & \multicolumn{3}{c}{5-way 1-shot} & \multicolumn{3}{c}{5-way 5-shots} \\
 &  & P & R & F1 & P & R & F1 \\
Model Type & Ablation &  &  &  &  &  &  \\
\midrule
\multirow[t]{4}{*}{Hybrid} & Original & 52.41 $\pm$ 4.07 & 33.03 $\pm$ 1.01 & 40.50 $\pm$ 1.97 & 55.31 $\pm$ 2.04 & 49.98 $\pm$ 2.55 & 52.50 $\pm$ 2.28 \\
 & No Paraphrases & 37.18 $\pm$ 2.80 & 29.30 $\pm$ 1.00 & 32.74 $\pm$ 1.46 & 42.55 $\pm$ 2.16 & 49.98 $\pm$ 2.34 & 45.95 $\pm$ 2.06 \\
 & No data pre-processing & 41.92 $\pm$ 3.18 & 22.48 $\pm$ 1.40 & 29.24 $\pm$ 1.79 & 54.70 $\pm$ 3.30 & 40.06 $\pm$ 2.84 & 46.18 $\pm$ 2.48 \\
 & No Rule/Sentence Augmentation & 46.59 $\pm$ 3.02 & 34.03 $\pm$ 1.37 & 39.32 $\pm$ 1.96 & 46.80 $\pm$ 1.79 & 53.56 $\pm$ 2.80 & 49.94 $\pm$ 2.09 \\
\midrule
\multirow[t]{4}{*}{SoftRules} & Original & 53.41 $\pm$ 4.25 & 32.54 $\pm$ 1.15 & 40.42 $\pm$ 2.10 & 59.38 $\pm$ 2.35 & 48.61 $\pm$ 2.61 & 53.45 $\pm$ 2.51 \\
 & No Paraphrases & 37.50 $\pm$ 2.86 & 28.87 $\pm$ 1.03 & 32.59 $\pm$ 1.51 & 44.60 $\pm$ 2.22 & 49.18 $\pm$ 2.47 & 46.77 $\pm$ 2.20 \\
 & No data pre-processing & 41.83 $\pm$ 3.54 & 21.99 $\pm$ 1.54 & 28.80 $\pm$ 1.99 & 57.24 $\pm$ 3.31 & 37.87 $\pm$ 2.63 & 45.58 $\pm$ 2.90 \\
 & No Rule/Sentence Augmentation & 47.28 $\pm$ 3.13 & 33.57 $\pm$ 1.52 & 39.26 $\pm$ 2.10 & 49.21 $\pm$ 1.74 & 52.27 $\pm$ 2.95 & 50.68 $\pm$ 2.19 \\
\bottomrule
\end{tabular}}
\caption{\footnotesize{Ablation results on the 5-way 1-shot and 5-way 5-shot on the development partition of the few-shot TACRED dataset. Each ablation condition is tested independently, with only one modification applied compared to the Original model.}} 
\vspace{-5mm}
\label{tab:ablation}
\end{table*}

Next, we analyze the contributions of each key component in our proposed method. We show the results of the ablation study in Table~\ref{tab:ablation}. The three components that we analyze are: 
{\flushleft (i)} The pre-processing of our training dataset, where we filter out duplicates and sub-sample very frequent rules and entities.

{\flushleft (ii)} The data augmentation, where we randomly replace the entities in the rule and in the sentence with synonyms. 
For example, a rule such as \texttt{[ne=per]+ <nsubj founded >nmod\_in [ne=org]+} becomes \texttt{[ne=human]+ <nsubj founded >nmod\_in [ne=company]+}. Similar augmentation are performed to sentences as well, where the named entity in the marker \cite{zhou2021improved} is changed with its synonyms.

{\flushleft (iii)} The inclusion of paraphrases. For example, a sentence such as \textit{Bill Gates founded Microsoft} can be automatically paraphrased into \textit{Bill Gates is the founder of Microsoft} using an LLM without losing any semantic information. 

The analysis in Table~\ref{tab:ablation} indicates that all three components contribute to the final performance, to varying degrees. 
First, our findings suggest that the data pre-processing contributes the most to the final performance, suggesting that the quality and structure of the input data play a crucial role in preparing the model to accurately handle the complexities of relation classification tasks. 
Second, the decline observed in the 'No paraphrases' setting suggests that the inclusion of paraphrases encourages the model to learn less obvious semantic variations.
Third, the rule and sentence augmentation appear to have the lowest impact. We argue that this is because both datasets that we use, the few-shot variants of TACRED and NYT29, contain the same common named entities, such as \texttt{person} and \texttt{organization}. These entities were seen during training, due to their prevalence. We hypothesize that this augmentation shines when the named entities used in the rules are not seen during training. We leave this exploration to future work.

\subsection{Are Soft Matching Rules still Pliable?}
One key advantage of rules is that they are pliable (see Footnote~\ref{fn:pliability}) and modular. This means that a domain expert is able to modify the model effectively without risking introducing unknown and undesirable behavior \cite{Sculley2015HiddenTD}.

We analyze the degree to which interventions on the resulting rules can improve the final performance. We choose the relation \texttt{org:parents} from the development set, as it is a relation relatively well represented in the dataset and one where our model obtains a lower F1 score. 
We design the following experiment: two experts have access to the syntactic rules associated with the support sentences from the development partition of the Few-Shot TACRED. They have up to two hours to improve the rule set and the following operations:

{\flushleft \textbf{ADD Rule:}} Adds a new rule which will be available to every episode. This operation simulates the practical example where practitioners aim to incorporate new knowledge to the model to be used at inference time.

{\flushleft \textbf{DELETE Rule:}} For a given support sentence with the relation \texttt{org:parents} in a given episode, the model will not have access to the rule generated on that support sentence.

{\flushleft \textbf{MODIFY Rule:}} This operation modifies a given rule. This modification will only be visible in the episodes for which this particular rule appears.

We show the statistics of the operations used by the experts in Appendix~\ref{a1:ss:stats}.

We show our results in Table~\ref{tab:tacred_human_interventions}. We detail two sets of results, showcasing the adaptability and effectiveness of our proposed method in relation classification. 
The first set is based on expert rule modifications without altering the classification threshold. The second set, in contrast, involves an increase in the threshold specifically for the \texttt{org:parents} rules, motivated by the greater average similarity seen with more general rules (created by the human annotators) compared to the lower alignment of highly specific rules (generated automatically from support sentences). 
For instance, rules synthesized from support sentences often yield highly specific constructs, such as \texttt{[ne=org]+ <nmod\_from taken >conj\_and operating >nmod\_under brandname >compound [ne=org]+}. Such rules typically align poorly with the majority of sentences, attracting lower similarity scores. In contrast, the introduction of more general rules, e.g.: \texttt{[ne=org]+ >appos subsidiary >nmod\_of [ne=org]+}, enhances rule-to-sentence similarity. This observed increase in average similarity was not accounted for with the original, unchanged classification threshold. To address this, we conducted a second set of experiments where the threshold was selectively increased by 0.1, but only for the \texttt{org:parents} relation. 

We observe a consistent performance increase across both expert interventions and both threshold scenarios. With the classification threshold held constant, expert modifications led to an improvement of approximately 4 F1 points, a relative increase of about 25\%. When the threshold for the \texttt{org:parents} relation was raised, the performance gains were even more pronounced, exceeding 15 F1 points and representing a relative increase of around 100\%. Notably, these enhancements did not adversely affect the performance on other relations.

\begin{table}[t]
\centering
\footnotesize
\resizebox{0.8\columnwidth}{!}{\begin{tabular}{@{\extracolsep{4pt}}lrr}
\toprule   
{Model} & \multicolumn{1}{c}{Original threshold}  & \multicolumn{1}{c}{Stricter threshold}\\
 
\midrule
Original     & 15.57 $\pm$ 1.39 & 15.57 $\pm$ 1.39 \\
\midrule
Expert 1     & 19.42 $\pm$ 0.65 & 31.78 $\pm$ 2.18 \\
Expert 2     & 19.77 $\pm$ 1.08 & 34.03 $\pm$ 1.91\\
\bottomrule
\end{tabular}

}
\caption{\footnotesize{F1 performance for the relation \texttt{org:parents} after two domain experts individually modified the corresponding rules. We compare scores before and after these changes, in two settings: (i) same threshold, (ii) stricter threshold.}}
\label{tab:tacred_human_interventions}
\vspace{-6mm}
\end{table}

\section{Conclusion}
We introduced a novel neuro-symbolic approach for relation extraction that combines the better generalization of neural networks with the explainability and pliability of rules. Our method first attempts to match the rule in a typical binary match/no match way. When a rule does not match, our approach then semantically matches it over text using a semantic matching component, which is contrastively trained without any human-annotated training data, akin to an LLM for rules.

We evaluated our model on two challenging few-shot datasets: Few-Shot TACRED \cite{sabo-etal-2021-revisiting} and a few-shot variant of NYT29 \cite{Fahmida2024MetaDatasetRE}. We showed that our method achieves strong performance, outperforming state-of-the-art supervised methods in three out of the four settings we investigated. Moreover, we empirically validated that our proposed method retains the pliability of rule-based methods, i.e., where humans can refine the underlying classification rules to noticeably increase the final performance. Notably, 
the resulting model is relatively small, i.e., it consists of an encoder of approximately 350M parameters, which makes it considerably more efficient than a decoder-based LLM.\footnote{Code available at \url{https://github.com/robertvacareanu/softrules}}
\section*{Acknowledgments}
\label{acknowledgments}

This work was partially supported by the
Defense Advanced Research Projects Agency (DARPA) under the ASKEM program. Mihai Surdeanu declares a financial interest in \url{lum.ai}. This interest has been properly disclosed to the University of Arizona Institutional Review Committee and is managed in accordance with its conflict of interest policies.

\section*{Limitations}
%
%

We evaluate our proposed approach only for the English language, where high-quality syntactic parsers are available, and relation classification, where most relations to be learned can be well covered by syntactic patterns. Nevertheless, thanks to efforts such as Universal Dependencies \cite{nivre-etal-2020-universal}, high-quality parsing data is available to a large number of languages. 

In general, rules seem to perform best for closed-world scenarios common to information extraction tasks. It is not immediately obvious how well rules (even with the proposed ``soft'' match) would port to more open-ended tasks such as question answering.

\section*{Ethics Statement}

Our approach uses pre-trained language models as the backbone of our soft matching component. Therefore this work shares many of the same ethical issues such as social biases or perpetuating stereotypes \cite{Weidinger2021EthicalAS}. 
Our work attempts to improve upon these by using a sieve architecture, where the first step is to apply the rule as in a typical rule-based model. This step is completely transparent to the practitioner, as they can add, modify, or delete rules. In the second step, we use a transformer-based model to semantically match the rules with sentences where an exact match is not possible.
Our pliability experiment showed that our approach retains the benefits of typical rule-based models, as the experts are able to intervene on the rules, and, thus, correct any potential biases that may exist. However, we acknowledge that more work is necessary to better understand the transparency of the semantic-matching component.

\bibliography{anthology,custom}

\appendix

\appendix

\newpage
\section{Qualitative Examples}
\label{appendix:qualitative_examples}
We provide qualitative examples for the behavior of our proposed semantic rule matcher (SRM) in Table~\ref{appendix:tab:qualitative_examples}.

We split the examples into 7 distinct blocks to facilitate the highlight of different behaviors. 

{\flushleft {\bf (1)}} In the first block we highlight how the SRM is able to overlook superficial differences (i.e. \textit{daughter} in text, \textit{son} in rule) and assign a high similarity score. We want to emphasize that a traditional rule-based engine will not be able to match the rule on the given sentence.

{\flushleft {\bf (2)}} Similar to block \textbf{(1)}, the SRM is capable of understanding that \textit{graduated from} is similar to \textit{got his degree from}.

{\flushleft {\bf (3, 4)}} We use these blocks to highlight to give a similarity reference for the behavior we want to highlight next. Here, the SRM assigns a high score, as expected. We want to highlight that this rule, in this form, is generic enough to match relations such as \texttt{neighborhood of}, \texttt{city in country}, among others.

{\flushleft {\bf (5, 6, 7, 8)}} In these blocks we highlight a behavior that is present in the resulting model, despite never being trained for it.
Here, we replace the typical named entities with their most fine-grained version: lexicalized entities. 
The underlying idea is to overcome the lack of expressiveness from the NER model and provide an additional source of signal, from the underlying entities.
In block \textbf{(5)} we replace the \texttt{location} entity types with \textit{Wynwood} and \textit{Miami}.\footnote{\textit{Wynwood} is a neighborhood in Miami.} 
We want to highlight that this rule correctly obtains a higher similarity with the sentence in block \textbf{(5)} than with the sentence in block \textbf{(6)}, where the entities in the sentence are \textit{Athens} and \textit{Greece}. We remark that the underlying relation in \textbf{(5)} is, in the most specific form, \textit{neighborhood of}, while in \textbf{(6)} it is \textit{city in country}. 
Similarly, we provide the alternative rule and the corresponding similarities in blocks \textbf{(7, 8)}. We emphasize that the SRM component has not been explicitly trained for this behavior.
We leverage this behavior during evaluation for the cases where both entity types were identical (e.g., \texttt{[ne=location]+ <appos [ne=location]+})

\begin{table*}
    \resizebox{\textwidth}{!}{\begin{tabular}{lll}
    \toprule
    \multirow{3}{*}{\Huge{1~~~~~}} &
    \textbf{\Large{Sentence}} & \begin{tabular}[c]{@{}l@{}}\Large{\underline{Sofia Coppola} , daughter of \underline{Francis Ford Coppola} , is one of the } \\ \Large{few to succeed in doing so : her film'' Lost in Translation'' won her a screenplay Oscar}\end{tabular} \\
    \addlinespace[0.25em]
    & \textbf{\Large{Rule}} & \Large{[ne=person]+ >appos son >nmod\_of [ne=person]+} \\
    \addlinespace[0.25em]
    & \textbf{\Large{Similarity}} & \Large{0.83} \\
    \addlinespace[0.25em]
    \midrule
    \multirow{3}{*}{\Huge{2}} &
    \textbf{\Large{Sentence}} & \Large{\underline{John} got his degree from \underline{Oxford} .} \\
    \addlinespace[0.25em]
    & \textbf{\Large{Rule}} & \Large{[ne=person]+ graduated from [ne=organization]+} \\
    \addlinespace[0.25em]
    & \textbf{\Large{Similarity}} & \Large{0.82} \\
    \addlinespace[0.25em]
    \midrule
    \multirow{3}{*}{\Huge{3}} &
    \textbf{\Large{Sentence}} & \Large{John moved to \underline{SoHo} , \underline{Manhattan} .} \\
    \addlinespace[0.25em]
    & \textbf{\Large{Rule}} & \Large{[ne=location]+ <appos [ne=location]+} \\
    \addlinespace[0.25em]
    & \textbf{\Large{Similarity}} & \Large{0.68} \\
    \addlinespace[0.25em]
    \midrule
    \multirow{3}{*}{\Huge{4}} &
    \textbf{\Large{Sentence}} & \Large{John moved to \underline{Athens} , \underline{Greece} .} \\
    \addlinespace[0.25em]
    & \textbf{\Large{Rule}} & \Large{[ne=location]+ <appos [ne=location]+} \\
    \addlinespace[0.25em]
    & \textbf{\Large{Similarity}} & \Large{0.69} \\
    \addlinespace[0.25em]
    \midrule
    \multirow{3}{*}{\Huge{5}} &
    \textbf{\Large{Sentence}} & \Large{John moved to \underline{SoHo} , \underline{Manhattan} .} \\
    \addlinespace[0.25em]
    & \textbf{\Large{Rule}} & \Large{[ne=Wynwood]+ <appos [ne=Miami]+} \\
    \addlinespace[0.25em]
    & \textbf{\Large{Similarity}} & \Large{0.29} \\
    \addlinespace[0.25em]
    \midrule
    \multirow{3}{*}{\Huge{6}} &
    \textbf{\Large{Sentence}} & \Large{John moved to \underline{Athens} , \underline{Greece} .} \\
    \addlinespace[0.25em]
    & \textbf{\Large{Rule}} & \Large{[ne=Wynwood]+ <appos [ne=Miami]+} \\
    \addlinespace[0.25em]
    & \textbf{\Large{Similarity}} & \Large{0.21} \\
    \addlinespace[0.25em]
    \midrule
    \multirow{3}{*}{\Huge{7}} &
    \textbf{\Large{Sentence}} & \Large{John moved to \underline{SoHo} , \underline{Manhattan} .} \\
    \addlinespace[0.25em]
    & \textbf{\Large{Rule}} & \Large{[ne=Berlin]+ <appos [ne=Germany]+} \\
    \addlinespace[0.25em]
    & \textbf{\Large{Similarity}} & \Large{0.24} \\
    \addlinespace[0.25em]
    \midrule
    \multirow{3}{*}{\Huge{8}} &
    \textbf{\Large{Sentence}} & \Large{John moved to \underline{Athens} , \underline{Greece} .} \\
    \addlinespace[0.25em]
    & \textbf{\Large{Rule}} & \Large{[ne=Berlin]+ <appos [ne=Germany]+} \\
    \addlinespace[0.25em]
    & \textbf{\Large{Similarity}} & \Large{0.37} \\
    \addlinespace[0.25em]
    \bottomrule
\end{tabular}}
      \caption{\footnotesize{Qualitative examples of our semantic rule matcher, split into 7 blocks to highlight different behaviors.}}
    \label{appendix:tab:qualitative_examples}
\end{table*}

\section{Entity Types in the Training Dataset}
\label{appendix:entity_types_training_dataset}

We used the following entity type pairs when constructing our dataset consisting of rule and sentence pairs: \texttt{[(ORGANIZATION, ORGANIZATION), (ORGANIZATION, PERSON), (ORGANIZATION, COUNTRY), (ORGANIZATION, CITY), (ORGANIZATION, STATE\_OR\_PROVINCE), (ORGANIZATION, IDEOLOGY), (ORGANIZATION, LOCATION), (ORGANIZATION, URL), (ORGANIZATION, EMAIL), (PERSON, ORGANIZATION), (PERSON, CAUSE\_OF\_DEATH), (PERSON, NATIONALITY), (PERSON, COUNTRY), (PERSON, LOCATION), (PERSON, CITY), (PERSON, STATE\_OR\_PROVINCE), (PERSON, IDEOLOGY), (PERSON, CRIMINAL\_CHARGE), (PERSON, RELIGION), (PERSON, EMAIL), (PERSON, MONEY), (TITLE, PERSON), (CITY, ORGANIZATION), (CITY, STATE\_OR\_PROVINCE), (PERSON, PERSON), (PERSON, TITLE), (PERSON, NUMBER), (COUNTRY, ORGANIZATION), (ORGANIZATION, COUNTRY), (NATIONALITY, PERSON), (PERSON, DATE), (COUNTRY, PERSON), (CITY, PERSON), (STATE\_OR\_PROVINCE, PERSON), (ORGANIZATION, DATE), (NUMBER, PERSON), (DATE, PERSON), (ORGANIZATION, NUMBER), (CAUSE\_OF\_DEATH, PERSON), (DATE, ORGANIZATION), (LOCATION, ORGANIZATION)]}.

\section{Entity Types Synonyms}
\label{appendix:entity_types_synonyms}
In the training phase of the proposed Semantic Rule Matcher, we randomly replaced the entity types in the rules and in the sentences with synonyms, to encourage generalization beyond superficial clues from the entity types. We present the synonyms we used in Table~\ref{appendix:tab:entiy_synonyms_table}.

\begin{table*}
    \resizebox{\textwidth}{!}{\begin{tabular}{@{}ll@{}}
\toprule
Entity & Synonyms \\
\midrule
organization & org, company, firm, corporation, enterprise \\
date & a specific date \\
person & per, human, human being, individual \\
number & digits \\
title & designation, formal designation \\
duration & time period \\
misc & miscellaneous \\
country & nation, state, territory \\
location & place, area, geographic area, loc \\
cause\_of\_death & date of demise, cause of death, death cause, mortal cause \\
city & municipality, town, populated urban area \\
nationality & citizenship \\
ordinal & ranking \\
state\_or\_province & region, territorial division within a country \\
percent & percentage \\
money & currency \\
set & collection, group of items \\
ideology & doctrine, system of ideas and ideals \\
criminal\_charge & accusation, formal allegation \\
time & period, time period \\
religion & belief, faith, spiritual belief, worshipper \\
url & web address \\
email & electronic mail \\
handle & username, personal identifier \\
\bottomrule
\end{tabular}
}
      \caption{Entity type synonyms used to augment the rules and sentences.}
    \label{appendix:tab:entiy_synonyms_table}
\end{table*}

\section{Paraphrasing Prompt}
\label{appendix:paraphrasing_prompt}
We show the prompt we used to generate paraphrases below. We dynamically set the number of paraphrases to generate based on the text length, ranging from $2$ to $5$. The intuition is that short sentences admit a lower number of paraphrases.
We only keep the paraphrases where the entities of interest are preserved. Additionally, if the entities of interest appear more than one time in the paraphrase, we discard the resulting paraphrase. 
Following this process, we keep over 80\% of the paraphrases that are generated.
\begin{quote}
Please generate a number of \{how many\} paraphrases for the following sentence. Please ensure the meaning and the message stays the same and these two entities are preserved in your generations: "\{entity 1\}", "\{entity 2\}". \\
Please be concise. \\
```\\
\{text\}\\
```\\
1. \\
\end{quote}

\section{Pliability Experiment}
\label{a1:s:pliability}

We show the number of operations employed by each Expert in Table~\ref{a:tab:intervention_exp_stats}.

\label{a1:ss:stats}
\begin{table}[ht]
\resizebox{\columnwidth}{!}{\begin{tabular}{@{\extracolsep{4pt}}lrrr}
\toprule   
{} & \multicolumn{3}{c}{Operations} \\
 \cmidrule{2-4} 
& \multicolumn{1}{c}{ADD} & \multicolumn{1}{c}{MODIFY} & \multicolumn{1}{c}{DELETE} \\ 
\midrule
Expert 1 & 12 & 6 & 16 \\
Expert 2 & 12 & 3 & 28 \\
\bottomrule
\end{tabular}}
\caption{The number of operations performed by each expert during the intervention experiment.}
\label{a:tab:intervention_exp_stats}
\end{table}

\section{Hyperparameters}
We experiment with multiple settings where we vary the learning rate, the projection dimensions, and the weight decay. This search involved under 20 runs. We show our hyperparameters in Table~\ref{appendix:tab:hyperparameters}. We use the development partition of Few-Shot TACRED for early stopping.

\begin{table}
    \resizebox{\columnwidth}{!}{\begin{tabular}{ll}
    \toprule
    \textbf{\Large{Rule Encoder LR}} & \Large{3e-5} \\
    \addlinespace[0.25em]
    \textbf{\Large{Sentence Encoder LR}} & \Large{1e-5} \\
    \addlinespace[0.25em]
    \textbf{\Large{Projections LR}} & \Large{1e-4} \\
    \addlinespace[0.25em]
    \textbf{\Large{Logit Scale LR}} & \Large{3e-4} \\
    \addlinespace[0.25em]
    \textbf{\Large{Train Batch Size}} & \Large{512} \\
    \addlinespace[0.25em]
    \textbf{\Large{Gradient Clip Val}} & \Large{5.0} \\
    \addlinespace[0.25em]
    \textbf{\Large{Dropout}} & \Large{0.1} \\
    \addlinespace[0.25em]
    \textbf{\Large{Projection Dims}} & \Large{384} \\
    \addlinespace[0.25em]
    \textbf{\Large{Weight Decay}} & \Large{0.001} \\
    \addlinespace[0.25em]
    \bottomrule
\end{tabular}}
      \caption{The hyperparameters we used for training the SRM.}
    \label{appendix:tab:hyperparameters}
\end{table}

\section{Hardware}
We ran all our experiments on a system with A100 80 GB GPUs. We used approximately 3 days worth of a single A100 GPU time.

\end{document}